\documentclass[sigconf]{acmart}

\AtBeginDocument{%
  \providecommand\BibTeX{{%
    \normalfont B\kern-0.5em{\scshape i\kern-0.25em b}\kern-0.8em\TeX}}}

\setcopyright{acmcopyright}
\copyrightyear{2018}
\acmYear{2018}
\acmDOI{10.1145/1122445.1122456}

\acmConference[Woodstock '18]{Woodstock '18: ACM Symposium on Neural
  Gaze Detection}{June 03--05, 2018}{Woodstock, NY}
\acmBooktitle{Woodstock '18: ACM Symposium on Neural Gaze Detection,
  June 03--05, 2018, Woodstock, NY}
\acmPrice{15.00}
\acmISBN{978-1-4503-XXXX-X/18/06}
\usepackage{amsmath}

\usepackage{graphics}
\usepackage{here}
\usepackage{color}

\usepackage{soul}
\sethlcolor{red}

\begin{document}

\title{SLGAN: Style- and Latent-guided Generative Adversarial Network for Desirable Makeup Transfer and Removal}

%

\author{Daichi Horita}
\affiliation{%
  \institution{The University of Tokyo}
}
\email{horita@hal.t.u-tokyo.ac.jp}

\author{Kiyoharu Aizawa}
\affiliation{
  \institution{The University of Tokyo}
}
\email{aizawa@hal.t.u-tokyo.ac.jp}



\begin{teaserfigure}
  \includegraphics[width=\textwidth,bb=0 0 2927 981]{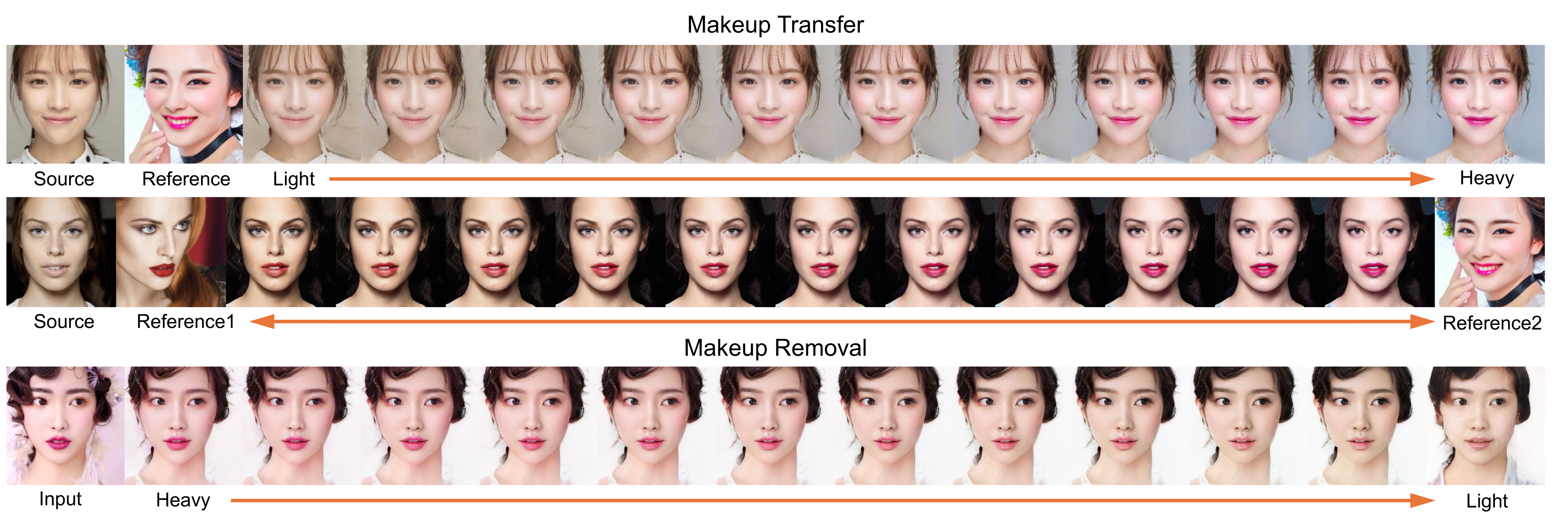}
  \caption{
    \textbf{Interpolation results of makeup transfer and removal.}
    We propose a style- and latent-guided generative adversarial network, which allows the user to adjust makeup shading in an image to obtain a desirable result.
    Our model interpolates from light to heavy makeup based on a style-guided value with a single reference image (first row) and two reference images (second row).
    Our model can also arbitrarily remove makeup by modulating a latent-guided value (third row).
  }
  \label{fig:teaser}
\end{teaserfigure}



%
%
\begin{abstract}
    There are five features to consider when using generative adversarial networks to apply makeup to photos of the human face.
    These features include (1) facial components, (2) interactive color adjustments, (3) makeup variations, (4) robustness to poses and expressions, and the (5) use of multiple reference images.
    Several related works have been proposed, mainly using generative adversarial networks (GAN).
    Unfortunately, none of them have addressed all five features simultaneously.
    This paper closes the gap with an innovative style- and latent-guided GAN (SLGAN).
    We provide a novel, perceptual makeup loss and a style-invariant decoder that can transfer makeup styles based on histogram matching to avoid the identity-shift problem.
    In our experiments, we show that our SLGAN is better than or comparable to state-of-the-art methods.
    Furthermore, we show that our proposal can interpolate facial makeup images to determine the unique features, compare existing methods, and help users find desirable makeup configurations.
    
\end{abstract}

\maketitle

\section{Introduction}

Many people are now using portrait-editing applications to transform their facial photos for experimental makeup presentations.
With extant tools, users can easily modify images using editing functions and perform trial-and-error procedures.
In the real world, makeup application is time consuming, it requires years of training, and one must maintain expert knowledge of products, colors, and application techniques.
Virtual makeup applications help alleviate this burden.
The YouCam Makeup virtual cosmetics application~\footnote{\url{https://www.perfectcorp.com/consumer/apps/ymk}} is a good example of this technology.
However, it remains difficult to find a virtual makeup application that always provides optimal suggestions to users.

Therefore, we have identified five features to consider when using generative adversarial networks (GANs) to apply makeup to photos of the human face.
These features include (1) facial components, (2) interactive color adjustments, (3) makeup variations, (4) robustness to poses and expressions, and (5) the use of multiple reference images.
Several studies of makeup transfer (MT) and removal (MR) have been proposed~\cite{paired_cycle_gan,beauty_gan,beauty_glow,psgan}, and most hove used GANs~\cite{gan,wgan}.
However, extant works have never striven to satisfy all five of the mentioned feature variables.
For example, BeautyGAN~\cite{beauty_gan} and PairedCycleGAN~\cite{paired_cycle_gan} could transfer and remove makeup.
However, they could not adjust the generated results.
BeautyGlow~\cite{beauty_glow} was limited to use of only one reference image.
PSGAN~\cite{psgan} was robust to changes of pose and expression, but it was not possible to perform MR.
In this paper, to adequately address all five features, we propose the SLGAN.
As shown in Figures~\ref{fig:teaser} and~\ref{fig:makeup_transfer_with_different_pose}, our framework effectively performs MT and MR while accounting for the five features mentioned above.

Our framework consists of a generator, a style encoder, a mapping network, and a discriminator as shown in Figure~\ref{fig:our_framework}. The generator comprises a shared encoder, a style-guided decoder, and a style-invariant decoder.
Our framework applies a strategy of scaling and shifting the generator parameters using the adaptive instance normalization (AdaIN)~\cite{adain,learned_representation}.
To achieve this strategy, we use a style encoder and a mapping network to obtain parameters $\beta$ and $\gamma$ for AdaIN.
As shown in Figure~\ref{fig:teaser}, this strategy enables SLGAN to perform the style- and latent-guided interpolations for the makeup adjustments.
Thus, users can adjust the generated results to find desirable makeup combinations with their reference images.
Furthermore, as shown in Figure~\ref{fig:makeup_transfer_with_different_pose}, our method is robust to poses and expressions.

The purpose of MT is to apply makeup while preserving the identity of the input image.
The style-guided decoder transfers makeup features, and the style-invariant decoder transfers plausible makeup features without using a style code.
Thus, we can tackle the problem of identity-shifting by computing the Euclidean distance between the outputs of the style-invariant and -guided decoders.
Moreover, we propose a novel perceptual makeup loss to help the generator apply more appropriate colors to the input.
The loss is used to compute a histogram of differences between the generated image and the reference image.
It thus enables our framework to adequately transfer makeup styles.

The major contributions of this paper are summarized as follows:

    \begin{enumerate}
        \item We propose a novel SLGAN framework for a MT and MR. This is the first style- and latent-guided framework for this task.
        \item Our proposed style-invariant decoder assists the generator to translate images that preserve the identity of the source.
        \item We propose a novel perceptual makeup loss that enables the generator to perform a high quality translation.
        \item Quantitative and qualitative experimental results show that SLGAN is better than or comparable to state-of-the-art methods. This framework enables the users to find the best makeup style by adjusting latent and style codes. 
        
    \end{enumerate}

\section{Related Works}
    \subsection{Style Transfer}
    
    Style transfer is the task of transferring a texture from a style image to a content image via an image translation.
    Gatys et al.~\cite{gatys_style_transfer} first demonstrated style transference by matching feature statistics in convolution layers of a deep neural network.
    Johnson et al.~\cite{Johnson2016Perceptual} proposed a perceptual loss function for training feed-forward networks for image translation tasks.
    Additionally, their proposed method achieved three orders-of-magnitude-faster inference times compared with the optimization-based method~\cite{gatys_style_transfer}.
    However, these methods are usually limited to a fixed set of styles and cannot adapt to arbitrary new styles.
    To solve this problem, Huang et al.~\cite{adain} proposed the AdaIN layer, which aligns the mean and variance of the content features with those of the style features.
    The style transfer applies a global style to an image, thus it is insufficient for a makeup transfer task.
    In this work, we instead propose an approach that transfers local styles to a local region of facial components.
s

\begin{figure*}[tb]
    \centering
    \includegraphics[width=\linewidth,bb=0 0 1253 466]{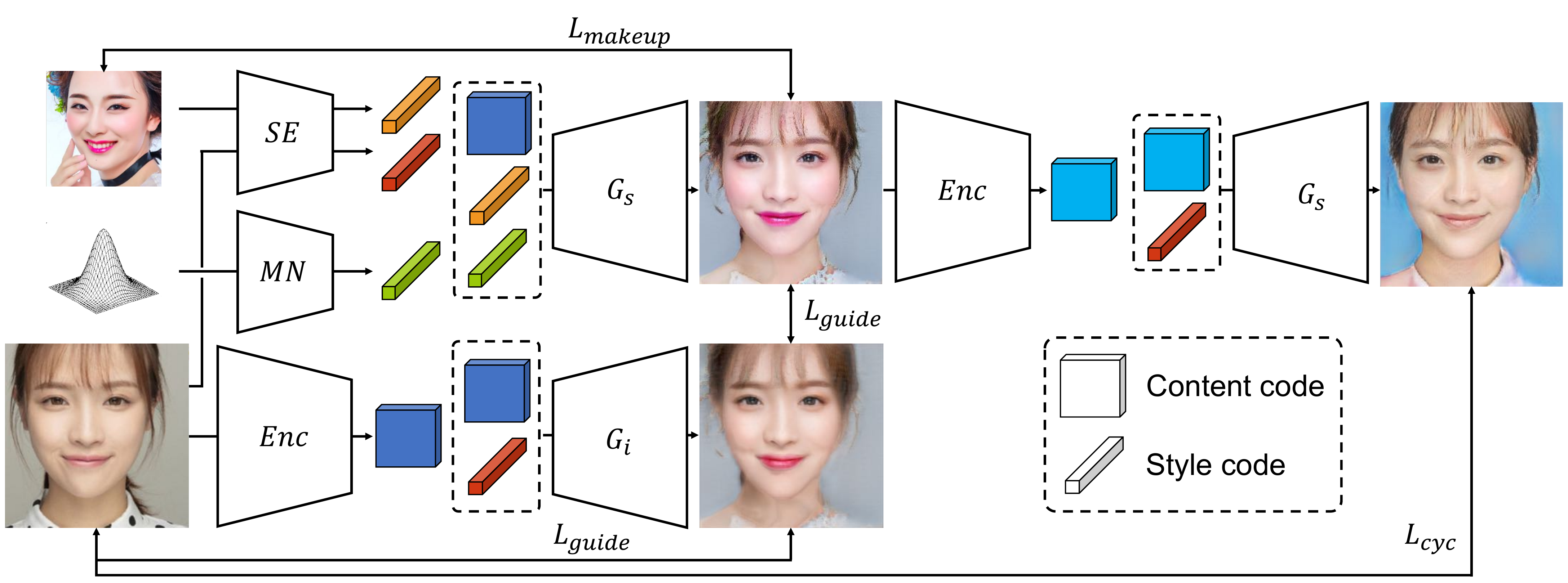}
    \caption{
        SLGAN consists of four modules: a generator $G$, a style encoder $SE$, and a mapping network $MN$. The generator $G$ consists of a shared encoder $Enc$, a style-guided decoder $G_{s}$, and a style-invariant decoder $G_{i}$.}
  \label{fig:our_framework}
\end{figure*}

\subsection{Guided Image-to-Image Translation}
    In a guided image-to-image translation task, given source and reference images, the goal is to train a network to translate an input image into its corresponding output image.
    There are two main approaches.
    The first is to use paired datasets~\cite{park2019spade,albahar2019guided,xian2017texturegan,liu2018patial_conv,li2017joint}.
    AlBahar et al.~\cite{albahar2019guided} proposed the use of spatially varying feature transformation and designed a bi-directional conditioning scheme that allowed the mutual modulation of the guidance and input network branches.

    The second approach is to leverage unpaired
    datasets~\cite{karras2018stylebased,starganv2,funit,munit,kim2019ugatit}.
    Karras et al.~\cite{karras2018stylebased} proposed a style-based generator based on a non-linear mapping network to embed the latent code in the style code.
    However, their proposed method did not employ an architecture to embed the reference image.
    FUNIT~\cite{funit} was a framework for few-show multi-domain unsupervised translation using reference samples from the target domain.
    StarGANv2~\cite{starganv2} provided both latent- and reference-guided synthesis and could be trained with coarsely labeled dataset.
    However, these methods often failed to transfer makeups, because they embedded global style features of the reference image were determined by the style code.
    As a result of global transfer, the problem of an identity-shift occurs, that is, the generated image loses the contents of the source image.
    Thus, these methods are not suitable for MT and MR
    Instead, we need an approach that can solve problems inherent in MT.

    To overcome this problem, out method encodes the makeup styles using a specific architecture for makeup problems.
    Furthermore, we introduce a style-invariant decoder to solve the identity-shift problem.

\subsection{Makeup Studies}
    The goal of MT is to perform style transfer based on semantic information while preserving the identity of the source image.
    PairedCycleGAN~\cite{paired_cycle_gan} proposed a method to train generators and discriminators for each face component.
    Given both source and reference images, BeautyGAN~\cite{beauty_gan} simultaneously trained MT and MR using a single generator and discriminator.
    Additionally, they proposed a makeup loss function, which matched the color histogram between the generated and reference images of facial components (e.g., lips, eye shadows, and whole face).
    Our style- and latent-guided framework differs from these architectures, and we consider the objective to be its optimization.
    We, therefore, propose a perceptual makeup loss that not only optimizes the network, but also encourages a multi-tasking learning.
    As a consequence, our network can learn to encode a reference image into higher quality style codes.

    LADN~\cite{ladn} proposed local adversarial discriminators to disentangle makeup features representations and contents to achieve local detail transference.
    However, this approach often failed to transfer in-the-wild images and could not partially adjust transfers.
    To overcome these problems, PSGAN~\cite{psgan} performed a MT using the Attentive Makeup Morphing module with an attention mechanism based on spatial information, using a style-guided architecture.
    Furthermore, PSGAN could adjust the proportion of the style of a reference image by adjusting the weight of attention features.
    However, PSGAN had a limitation with which it could not perform MR.
    In contrast, we apply an AdaIN with not only the style-guided architecture, but also with the latent-guided one to adjust makeup features.
    Thus, our model performs both MT and MR.
    As a result, our framework has more application-rich features.

\section{SLGAN}
    In this section, we introduce the details of our proposed method.
    First, in Section~\ref{sec:formulation}, we formulate the problem of style- and latent-guided MT and MR.
    In Section~\ref{sec:our_network}, we describe our proposed SLGAN architecture.
    Additionally, in Section~\ref{sec:style_invariant_guide_decoder}, we introduce a style-invariant decoder that assists the generator in the transference of makeup without losing any local face features.
    Then, in Section ~\ref{sec:perceptual_makeup_loss}, we present our proposed novel perceptual makeup loss, which enables a generator to reflect the color of the reference image, including lips and eye shadows.
    Finally, in Section~\ref{sec:other_loss}, we describe the other loss functions used to train our network.

\subsection{Formulation}\label{sec:formulation}
    Our goal is to extract makeup styles from the reference images and transfer them to the source images.
    Note that we consider transfers between the same class (e.g., from one makeup image to another).
    Let $X \subset \mathbb{R}^{H \times W \times 3}$ and $Y \subset \mathbb{R}^{H \times W \times 3}$ be the sets of the source and reference domains, where $H$ and $W$ represent the height and the width of input images, respectively.
    Let $Z \subset \mathbb{R}^{16}$ be the latent space.
    Additionally, we have $I_{s}^{X} \in X$ to represent source samples, $I_{r}^{Y} \in Y$ to represent reference samples, and $z \in Z$ to represent latent codes.
    Note that $X$ and $Y$ are the unpaired datasets.
    Thus, the problem can be formulated as one of unsupervised image translation conditioned by the reference image.
    That is, the source and reference images have different identities.
    Let $C \subset \mathbb{R}^{2}$ be the sets of MT-class conditions, including MR.
    We utilize $c \in C$ to represent makeup-transfer conditions.

\subsection{Network Architecture}\label{sec:our_network}
    \textbf{Overall}
    As shown in Figure~\ref{fig:our_framework}, we propose a style- and latent-guided framework for MT and MR.
    First, the style codes $s_{e}, s_{m} \in W$ are generated by a style encoder $SE$ and a non-linear mapping network $MN$ from the reference image $I_{r}^{Y} \in Y$ and the latent code $z \in Z$, respectively.
    Then, given an embedded style code $s$ and a source image $I_{s}^{X}$, our goal is to learn a generator $G: I_{s}^{X}, s \rightarrow \tilde I_{r}^{X}$, which transfers the style of the reference to the source.
    Note that the generator $G$ simultaneously learns the MT and MR.

    \textbf{Style encoder.}
    Given a reference image $I_{r}^{Y}$ and a one-hot vector $c$ representing its domain, the style encoder $SE$ learns to embed the reference image into a style code $s_{e}$, denoted as $SE_{c}: I_{r}^{Y}, c \rightarrow s_{e} \in W$.
    Our style encoder $SE$ is implemented using shared convolution layers and one-layer multi-layer perceptron (MLP).
    The style encoder $SE$ extracts features using an encoder and then applies the MLP layers per domain based on the control of the one-hot vector $c$.
    Therefore, because the style encoder uses each MLP layer for MT and MR, it can embed reference images in style codes with domain-specific representations.
    Note that, when the style encoder is given a reference image, a semantic mask is applied to remove the hair and background, which is generated by a face parsing algorithm~\footnote{\url{https://github.com/zllrunning/face-parsing.PyTorch}}.

    \textbf{Mapping network.}
    Given a latent code $z$ in the input latent space $Z$ and a random one-hot vector $c$, our non-linear mapping network $MN$ learns to embed a latent code in the style code $s_{m}$, denoted as $MN_{c}: z, c \rightarrow s_{m} \in W$.
    Our mapping network $MN$ is implemented using a shared six-layer MLP and an unshared one-layer MLP.
    Unlike a mapping network of StyleGAN~\cite{karras2018stylebased}, our mapping network enables the generator $G$ to generate delicate makeups.
    Additionally, our mapping network also yields a domain-specific style code $s_{m}$.
    Unlike a mapping network of StyleGAN~\cite{karras2018stylebased}, our mapping network enables the generator, $G$, to generate delicate

    \textbf{Adaptive normalization layer.}
    We use AdaIN~\cite{adain} with the style-guided decoder to perform MT and MR based on the style codes $s_{e}$ and $s_{m}$ of the reference image $I_{r}^{Y}$ and the latent code $z$, respectively.
    The style codes $s_{e}$ and $s_{m}$ control $\beta$ and $\gamma$ in the AdaIN operation after each convolution layer of the generator $G$.
    Note that these parameters are not per-parts of the face, but they are the features of the whole face.
    As a result, Our framework can perform a partial MT using multiple person's face parts.
    Then, the features of each source image $I_{s}^{X}$ are individually normalized and the scaling and shifting operations are performed using scalar components based on the style codes $s$.

    \textbf{Generator.}
    Given a source image $I_{s}^{X}$ and a style code $s$, our generator generates an image $\tilde I_{r}^{X}$, that preserves both the makeup style of the reference image $I_{r}^{Y}$, and the identity of the source $I_{s}^{X}$.
    Our generator is implemented in an encoder--bottleneck--decoder~\cite{Johnson2016Perceptual,stargan} network.
    As shown in Figure~\ref{fig:our_framework}, our generator $G$ consists of a shared encoder $Enc$, a style-guided decoder $G_{s}$, and a style-invariant decoder $G_{i}$.
    To simplify the notation, we denote a style-guided generator as $G_{sg}(I_{s}^{X}, s) = G_{s}(Enc(I_{s}^{X}), s)$ and a style-invariant generator as $G_{ig}(I_{s}^{X}) = G_{i}(Enc(I_{s}^{X}))$.
    Each decoder has the same structure except for the normalization layer.
    A shared encoder $Enc$ embeds a source image $I_{s}^{X}$ in a content code.
    A shared encoder and a style-invariant decoder have an instance normalization~\cite{instance_norm} so that they can make the features conform to a normal distribution.
    An adaptive wing based heatmap~\cite{adaptive_wing_heatmap} of a source image $I_{s}^{X}$ is added to each feature map.

    \textbf{Discriminator.}
    To make the generator $G$ generate realistic images, we use the discriminator $D$.
    Our discriminator $D$ has the same structure as the style encoder $SE$.
    Additionally, it is a multi-task discriminator~\cite{funit,mescheder2018training,starganv2} that has multiple linear output branches.
    Therefore, each branch learns domain-specific features.

\subsection{Style-invariant Decoder}\label{sec:style_invariant_guide_decoder}
    There is an identity-shift problem in which the generator cannot preserve the identity of the source image when the global style of the reference image is embedded in the style code.
    Thus, the discrepancy of identities between reference and source images can cause problems in which the generated image cannot maintain the content of the source.
    To overcome this problem, our proposed style-invariant decoder generates images from the shared feature without the style code, which is extracted by a shared encoder.
    That is, this network has no AdaIN layers.
    In the field of coloring sketches, some studies~\cite{tag2pix,guide_decoder} have used a guide decoder to avoid the gradient disappearance in mid-level layers.
    On the other hand, our style-invariant decoder helps the generator not only perform stable learning like the guide decoder, but it also helps avoid an identity-shift problem.
    The style-invariant decoder is only used for training, not for testing.

\subsection{Perceptual Makeup Loss}\label{sec:perceptual_makeup_loss}
    To further encourage the network to transfer makeup per face component, a constraint on the consistency of makeup styles should be accounted for in the network.
    To satisfy this constraint, based on the observation that face makeup  is considered a color distribution~\cite{beauty_gan}, we propose a new histogram-matching strategy and propose perceptual makeup loss.
    In our framework, the key idea is for the style encoder to have a structure for extracting makeup and non-makeup styles.
    The perceptual makeup loss computes the histogram matching using features of each convolution layer of a style encoder between the generated image and the reference image.
    This encourages the style encoder to learn better parameters through a multi-task learning.
    This loss function entails the integration of three local histogram losses acting on the lips, eyes, and facial regions, defined as
    \begin{eqnarray}
        \label{eq:new_makeup_loss_1}
        \mathcal{L}_{makeup} = \lambda_{lips} \mathcal{L}_{lips} + \lambda_{eyes} \mathcal{L}_{eyes} + \lambda_{face} \mathcal{L}_{face},
    \end{eqnarray}
    \begin{eqnarray}
        \label{eq:old_makeup_loss_1}
        \mathcal{L}_{item} = \sum_{l=1}^{K} || \phi_{l}(\tilde I_{r}^{X}) - HM(\phi_{l}(\tilde I_{r}^{X} \circ S_{item}^{1}), \phi_{l}(I_{r}^{Y} \circ S_{item}^{2}) ) ||_{2},
    \end{eqnarray}
    \begin{eqnarray}
        \label{eq:old_makeup_loss_2}
        S_{item}^{1} = FP(\tilde I_{r}^{X}), S_{item}^{2} = FP(I_{r}^{Y}),
    \end{eqnarray}
    where $\phi_{l}$ denotes a $l$-th layer feature map, $K$ denotes the sum of the number of convolution layers, $\circ$ denotes element-wise multiplication, $item$ denotes the set of \{$lips, eyes, face$\}, $FP$ denotes the face parsing algorithm, $HM$ denotes the histogram matching operation, and $S$ denotes the semantic mask of face components.
    In Section~\ref{sec:ablation}, we describe the suitability of the style encoder to obtain the feature maps.

\subsection{Other Objectives}\label{sec:other_loss}
    Additionally, regarding the perceptual makeup loss described in Section~\ref{sec:perceptual_makeup_loss},
    we use the following objectives, which are similar to related works~\cite{starganv2,mao2019mode,yang2019diversitysensitive,CycleGAN}.

    \textbf{Adversarial Loss.}
    To make the generated images more realistic, we adopt an adversarial loss, defined as
    \begin{eqnarray}
            \label{loss:adversarial}
            \mathcal{L}_{adv} &=&\underset{G_{s}}{\rm{min}}~ \underset{D_{c}}{\rm{max}}
             ~ {\mathbb{E}_{I_{s}^{X}, c}} \left[\log{D_{c}(I_{s}^{X})}\right]+ \nonumber \\
          && \mathbb{E}_{I_{s}^{X}, \hat{c}, \hat{s}}
          [\log{(1 - D_{\hat{c}}(G_{sg}(I_{s}^{X}, \hat{s})))}],
    \end{eqnarray}
    where the target style code $\hat{s}$ is generated by a style encoder $\hat{s_{e}} = SE_{\hat{c}}(I_{r}^{Y})$ and a non-linear mapping network $\hat{s_{m}} = MN_{\hat{c}}(z/)$.
    $c$ and $\hat{c}$ represent the source domain and target domain, respectively.
    $D_{c}$ represents the corresponding domain of $c$ and $G_{sg}$ represents the style-guided generator.
    A discriminator distinguishes whether the generated image $\tilde I_{r}^{X}$ is a real or not.
    
    \textbf{Style diversity loss.}
    We introduce a regularization term to spread over the generated space~\cite{mao2019mode,yang2019diversitysensitive}, which is defined as
    \begin{eqnarray}
            \label{loss:style_variegation}
            \mathcal{L}_{sd} &=&{\mathbb{E}_{I_{s}^{X},\hat{c}, z_{1}, z_{2}}} \left[ || G_{sg}(I_{s}^{X}, \hat{s_{1}})) - G_{sg}(I_{s}^{X}, \hat{s_{2}}) ||_{1} \right],
    \end{eqnarray}
    where $\hat{s_{1}}$ and $\hat{s_{2}}$ are generated by a style encoder $SE_{\hat{c}}$ or a mapping network $MN_{\hat{c}}$ from random latent codes $z_{1}$ and $z_{2}$, and a target condition vector $\hat{c}$, denoted as $\hat{s_{e}} = SE_{\hat{c}}(z)$ and $\hat{s_{m}} = MN_{\hat{c}}(z)$, respectively.
    This encourages the generator to explore the latent code and increases the chance of generating various samples.
    The discriminator learns better parameters, because it properly classifies samples that are rarely generated.
    As a result, by using this objective, our framework properly learns fine makeup styles.

    \textbf{Style reconstruction loss.}
    To constrain the style codes to correctly represent the style of makeup or non-makeup, we use the style reconstruction loss~\cite{munit,zhu2017multimodal}, defined as 
    \begin{eqnarray}
        \label{loss:style_identity}
        \mathcal{L}_{sr} &=&{\mathbb{E}_{I_{s}^{X}, \hat{c}, z}} \left[ || \hat{s} - SE_{\hat{c}}(G_{sg}(I_{s}^{X}, \hat{s})) ||_{1} \right].
    \end{eqnarray}
    This objective is similar to a latent reconstruction loss~\cite{chen2016infogan,zhang2016stackgan}.

    \textbf{Cycle consistency loss.}
    By optimizing Eq.(\ref{loss:adversarial},\ref{loss:style_variegation},\ref{loss:style_identity}), the generator can generate diverse and realistic images.
    However, the generator should not only preserve the features of the source image, but it should also fool the discriminator.
    As a result, there is a problem in which only these objectives do not guarantee that the generated image preserves the content of the source image.
    To solve this problem, we use the cycle consistency loss~\cite{CycleGAN,drit}, defined as
    \begin{eqnarray}
        \label{loss:cycle_consistency}
        \mathcal{L}_{cyc} &=&{\mathbb{E}_{I_{s}^{X},c,\hat{c},s}} \left[ || I_{s}^{X} - G_{sg}(G_{sg}(I_{s}^{X}, \hat s), \bar s) ||_{1} \right],
    \end{eqnarray}
    where $\hat s$ represents the style code of a target domain $\hat{c}$ and $\bar s$ represents an original domain $c$ of $I_{s}^{X}$, denoted as $\bar s = SE_{c}(I_{s}^{X})$.
    Minimizing this objective enables the generator to perform a MR and MR while preserving the contents of the source image.

    \textbf{Style-invariant guide loss.}
    Despite the use of cycle consistency loss, the generated image changes the shape of facial components, depending on makeup and non-makeup styles, owing the identity-shift problem.
    To achieve this problem, we propose a style-invariant guide loss to encourage the generated image to naturally apply the style of the reference image and maintain the content of the source image.
    Is is defined as
    \begin{eqnarray}
        \label{loss:new_guide}
        \mathcal{L}_{guide} &=&{\mathbb{E}_{I_{s}^{X}}} \left[ \lambda_\gamma || I_{s}^{X} - G_{ig}(I_{s}^{X}) ||_{2} \right] + \nonumber \\
        && {\mathbb{E}_{I_{s}^{X},\hat{c},s}} \left[ \lambda_\beta || G_{ig}(I_{s}^{X}) - G_{sg}(I_{s}^{X}, \hat{s}) ||_{2} \right],
    \end{eqnarray}
    where each $\lambda$ are hyper-parameters, $G_{ig}$ represents the style-invariant generator, $G_{sg}$ represents the style-guided generator, and $\hat s$ represents the style code of the target domain $\hat c$.
    We do not give the style code $\hat{s}$ to the style-invariant generator $G_{ig}$.

    \textbf{Total Loss.}
    Finally, the loss functions of $G$, $SE$, $MN$, and $D$, which are optimized in our framework, are defined as
    \begin{eqnarray}
        \label{loss:D}
        \mathcal{L}_{D} &=& - \lambda_{adv} \mathcal{L}_{adv}
    \end{eqnarray}
    \begin{eqnarray}
        \label{loss:G}
        \mathcal{L}_{G} &=& \lambda_{adv} \mathcal{L}_{adv} + \lambda_{sd} \mathcal{L}_{sd} + \lambda_{sr} \mathcal{L}_{sr} + \lambda_{cyc} \mathcal{L}_{cyc} \nonumber \\
        && + \lambda_{makeup} \mathcal{L}_{makeup} + \lambda_{guide} \mathcal{L}_{guide},
    \end{eqnarray}
    where each $\lambda$ is a hyper-parameter.
    
\begin{figure*}[t]
    \centering
    \includegraphics[width=\linewidth,bb=0 0 2311 588]{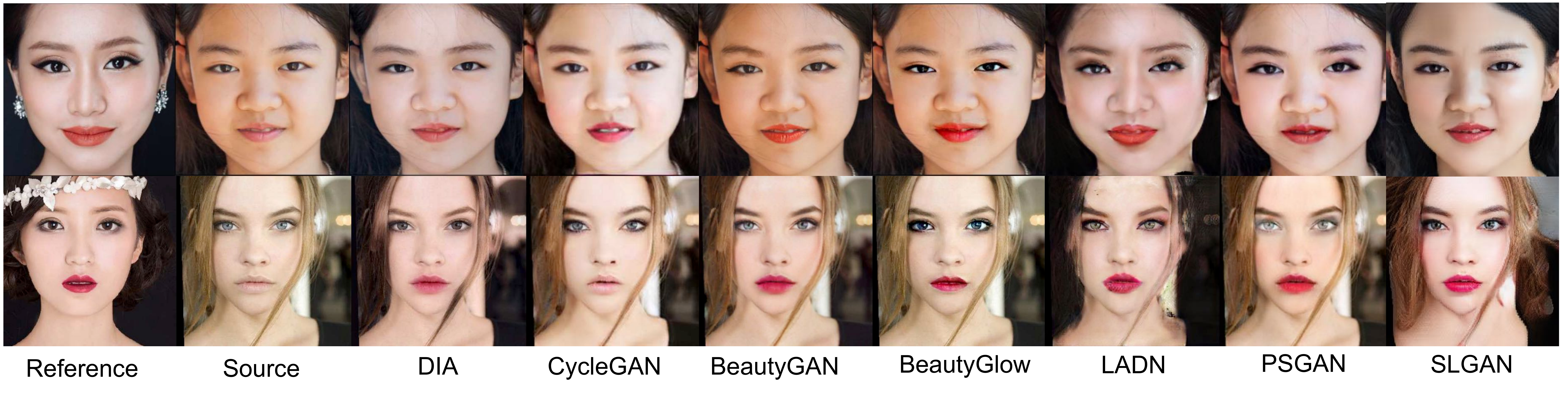}
    \caption{
         Qualitative comparison of makeup transfer of baseline methods and our style-guided SLGAN.
         Our method can generate images that are closer to the reference image from the views of lips, eyes, eye shadows, and skin tones.
    }
  \label{fig:makeup_transfer}
\end{figure*}

\section{Experiments}

In this section, we introduce details of our implementations and evaluations.
First, in Section~\ref{subsec:imple}, we provide the dataset and hyper-parameters.
Next, in Section~\ref{subsec:exp_makeup_transfer} and~\ref{subsec:exp_makeup_removal}, we conduct the qualitative and quantitative experiments of MT and MR and compare results with baselines.
Then, in Section~\ref{sec:ablation}, we conduct ablation studies to validate the effectiveness of our proposed methods.
Finally, in Section~\ref{subsec:interpolation}, we present the interpolation results of style- and latent-guided methods.

\subsection{Implementation Details}\label{subsec:imple}
\textbf{Dataset.}
    We use the Makeup Transfer (MT) dataset~\footnote{\url{http://liusi-group.com/projects/BeautyGAN}} provided by Li et al.~\cite{beauty_gan} for MT and MR.
    The dataset contains 3,834 facial images with a resolution of $256 \times 256$, consisting of 1,115 non-makeup images and 2,719 makeup unpaired images.
    The dataset includes some variations on race, pose, expression, and backgrounds.
    All faces are calibrated to the front, and they include various makeup styles, such as ``smoky-eyes'', ``flashy'', ``retro'', ``Korean'', and ``Japanese''.
    The test set consists of 100 and 250 non-makeup and makeup images, respectively.

\textbf{Training Details.}
    As shown in Figure~\ref{fig:our_framework}, SLGAN comprises a discriminator, a style encoder, a mapping network, and a generator.
    The generator has skip connections that are useful for identity mapping~\cite{he2016identity}.
    Given reference and source images, our style-guided SLGAN generates an image using the style encoder.
    Given the latent code and a source image, our latent-guided SLGAN generates an image via the mapping network.
    The MT dataset applies a $256 \times 256$ resolution.
    Our implementation uses PyTorch~\cite{pytorch}, and the training time is set to 4 days using one TITAN RTX graphics card.
    We set the batch size to four, owing to GPU memory limitations.
    We adopt the Adam optimizer~\cite{kingma2014adam} with $\beta_{1}$=0.0, $\beta_{2}$=0.99, and weight decay=$10^{-4}$.
    We set a learning rate of $10^{-4}$ for $G$, $SE$, and $D$, and $10^{-6}$ for $MN$.
    For testing, we manage the parameters of $G$, $SE$, and $MN$ using an exponential moving average~\cite{PGGAN,gan_exponential_moving}.
    The overall network is initialized using the He initialization~\cite{he2015delving}.
    Finally, we set $\lambda_{adv} = 1, \lambda_{sd} = -1, \lambda_{sr} = 1, \lambda_{lips}=10, \lambda_{eyes}=10, \lambda_{face}=0.1, \lambda_{gamma}=0.5$, and $\lambda_{beta}=0.5$ as the hyper-parameters.

\subsection{Makeup Transfer Results}\label{subsec:exp_makeup_transfer}

\begin{figure}[t]
    \centering
    \includegraphics[width=\linewidth,bb=0 0 1535 326]{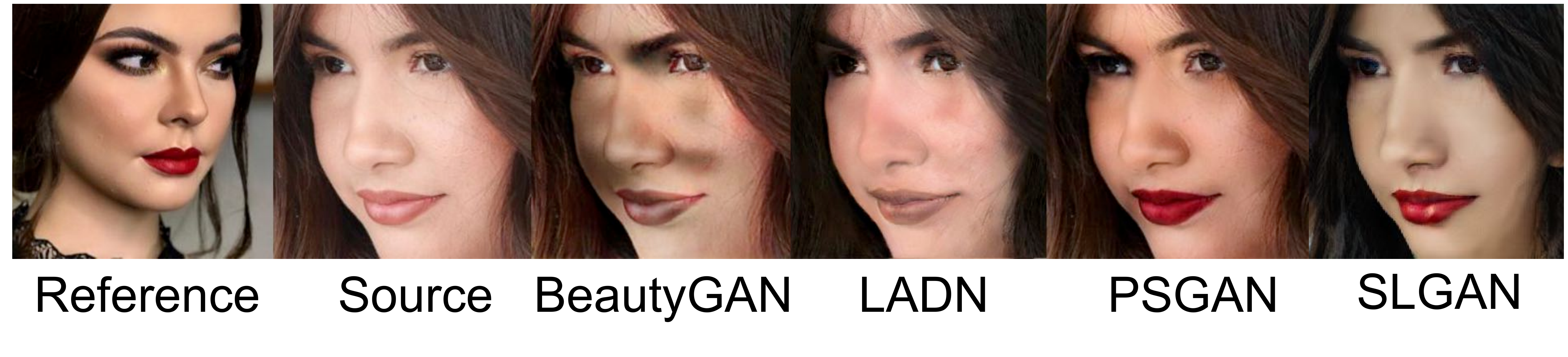}
    \caption{
        Qualitative comparison of makeup transfer with the source and reference images having different poses.
    }
  \label{fig:makeup_transfer_with_different_pose}
\end{figure}


We compared our style-guided SLGAN with baseline models on a MT task.
We did not use our latent-guided SLGAN, because we could not provide a reference image.
As baseline methods, we adopted two general image-to-image translation methods: DIA~\cite{dia} and CycleGAN~\cite{CycleGAN}.
We adopted five MT methods: BeautyGAN~\cite{beauty_gan}, PairedCycleGAN~\cite{paired_cycle_gan}, BeautyGlow~\cite{beauty_glow}, LADN~\cite{ladn}, and PSGAN~\cite{psgan}.

\textbf{Qualitative Comparison.}
Figure~\ref{fig:makeup_transfer} shows qualitative comparisons of SLGAN with the baseline methods.
Because the implementations of PSGAN and BeautyGlow are not publicly available, images were taken from their corresponding papers 
We observed that DIA failed to transfer makeup for the eyebrows, because it could not handle local regions.
CycleGAN demonstrated better MT for the eyebrows, compared with DIA.
However, it failed to transfer lip color.
Additionally, when using these general image-to-image translation methods, we could not edit the generated images.
The other MT baseline methods failed to transfer the pupil color of the reference image.
We argue that these capabilities are important, because people often use colored contact lenses to change their eye colors.

As shown in the lower row, BeautyGlow generated an image in which the eye shadow was clearly darker than that of the reference image.
LADN generated an image containing artifacts around the hair and barely retained the identity of the source image.
In the upper row, it can be seen that the PSGAN generated an unnatural results around the eyes. 
In the lower image, it can be seen that it also generated unnatural results that preserved the eye color of the source image and mixed the color features of the pupils with the eye shadows.
Compared with the baseline methods, our style-guided SLGAN generated images that were closer to the reference image based on the MTs of lips, eyes, eye shadow, and skin tones.

Figure~\ref{fig:makeup_transfer_with_different_pose} shows the results of different poses of source and reference images.
For the baseline, we used BeautyGAN~\cite{beauty_gan}, LADN~\cite{ladn}, and PSGAN~\cite{psgan}.
Note that we took the generated images of each baseline from the PSGAN paper. 
PSGAN trains its network using not only the MT dataset, but also their proposed Makeup-Wild dataset~\cite{psgan}, which contains images having a diversity of poses and facial expressions.
It is not publicly available, but we only used it to train our network.
From the comparisons, BeautyGAN and LADN failed to transfer makeup or generate artifacts. 
These methods did not provide an explicit structure for learning MT locations, and they overfitted the MT dataset, which contained only frontal images.
In contrast, SLGAN succeeded in transferring makeup, even without using the Makeup-Wild dataset.
Our framework learned the relationships between each face part, because the perceptual makeup loss was computed between the features of our style encoder.
Without the makeup loss, we observed a failure to learn relationships, as shown in Figure~\ref{fig:comparison_ablation_makeup_loss} (d).
In Figure~\ref{fig:comparison_ablation_makeup_loss} (a) and (c), we found that there was a failure of eye shadow and cheek makeup without perceptual makeup loss.
Additionally, a comparison between Figure~\ref{fig:comparison_ablation_guide_decoder} (a) and (c) shows that the style-invariant decoder assisted in transferring the makeup of the reference image.

Thus, with these two proposed modules, we can achieve its robustness.

\textbf{Quantitative Comparison.}
To provide quantitative evaluations of the MT, we conducted a user study using Amazon Mechanical Turk (AMT) in which 10 people participated.
We used CycleGAN~\cite{CycleGAN}, PairedCycleGAN~\cite{paired_cycle_gan}, BeautyGAN~\cite{beauty_gan}, and LADN~\cite{ladn} as baseline methods.
Given each generated image, a corresponding source, and a reference image, the Turkers were instructed to choose an image for which they felt the most natural makeup had been applied based on the reference image.
For a fair comparison, we shuffled the order of the generated images for each question.
We randomly selected 20 source and 20 reference images from the test set and generated images for all combinations.
From the generated results, we randomly selected 50 images per method.
Our latent-guided SLGAN could not be given reference images.
Thus, we did not use this method.
Table~\ref{table:qualitative_comparison} shows the results of the 10-person user study.
In this small scale experiment, our style-guided SLGAN had a better score, compared with the other methods.

\begin{figure}[t]
    \centering
    \includegraphics[width=\linewidth,bb=0 0 1840 1156]{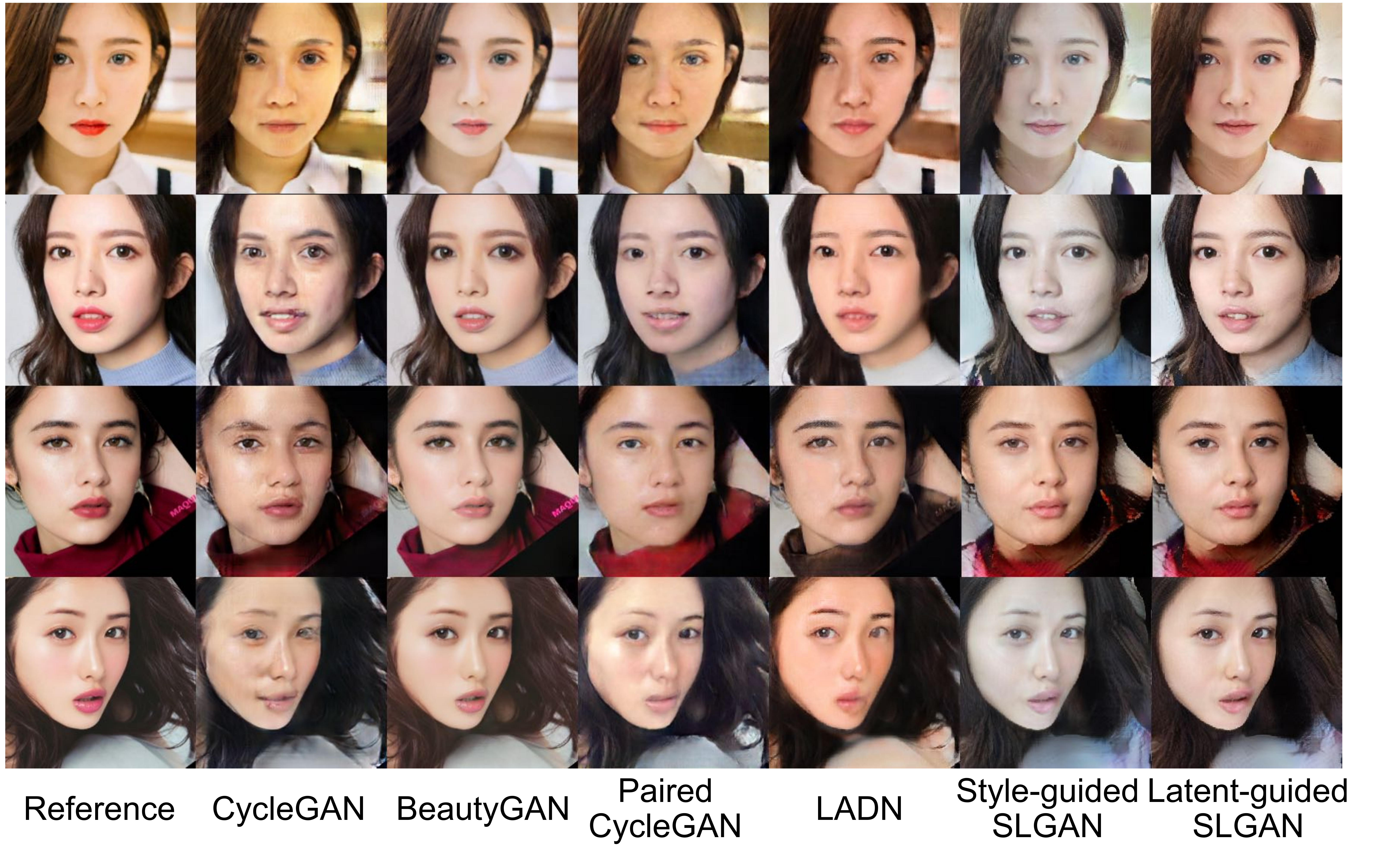}
    \caption{
        Qualitative comparison of makeup removal.
        Our style-guided and latent-guided generator performed makeup removal better than other methods.}
  \label{fig:makeup_removal}
\end{figure}

\begin{table}[t]
\caption{AMT evaluation for comparing baseline methods with SLGAN on tasks of a makeup transfer and removal.
Larger values indicate better performance.}
\label{table:qualitative_comparison}
\centering
\begin{tabular}{l|c|c}
\hline
Method             & Transfer (\%) & Removal (\%) \\ \hline
CycleGAN           & 17.2     & 5.6     \\
PairedCycleGAN     & 19.2     & 13.2    \\
BeautyGAN          & 22.0     & 1.8     \\
LADN               & 18.4     & 0.4     \\ \hline
Style-guided SLGAN  & \textbf{23.2}     & 38.6    \\
Latent-guided SLGAN & -        & \textbf{40.4}    \\ \hline
\end{tabular}
\end{table}

\subsection{Makeup Removal Results}\label{subsec:exp_makeup_removal}

We compared the proposed methods to the baseline models for a MR task.
We employed CycleGAN~\cite{CycleGAN}, BeautyGAN~\cite{beauty_gan}, PairedCycleGAN~\cite{paired_cycle_gan}, and LADN~\cite{ladn} as baseline methods.
We used our style- and latent-guided SLGAN.
Given a random choice of a non-makeup reference and a makeup image, our style-guided SLGAN translated the makeup image into a MR image using the style encoder to obtain a style code.
Given a randomly latent code and a makeup image, our latent-guided SLGAN translated a makeup image into a MR image using a mapping network to obtain the style code.

\textbf{Qualitative Comparison.}
Figure~\ref{fig:makeup_removal} shows qualitative comparisons between our SLGAN and other baseline methods.
CycleGAN showed a blurred image of poor quality.
Given a random non-makeup image, BeautyGAN performed decently, but it could not perform MR.
However, the method does not show images with MR.
Although PairedCycleGAN and LADN tended to remove makeup, they failed to generate clear lips and eyes.
In contrast, we found that our method produced clear MR images.
We observed that the images generated by our style-guided SLGAN were affected by the skin color of the given reference image.

\textbf{Qualitative Comparison.}
We conducted a user study of 10 people using AMT in the same setting as that of the MT.
As seen in Table~\ref{table:qualitative_comparison}, both our style- and latent-guided SLGAN showed better results compared with the baseline methods.
We can see that our style- and latent-guided SLGAN demonstrated similar quality MR with few differences.
We consider that our style-guided SLGAN performed MR based upon the skin color of the reference image, and it, therefore, scored lower than our latent-guided SLGAN.

\subsection{Ablation Study}\label{sec:ablation}
\begin{figure}[t]
    \centering
    \includegraphics[width=\linewidth,bb=0 0 859 1036]{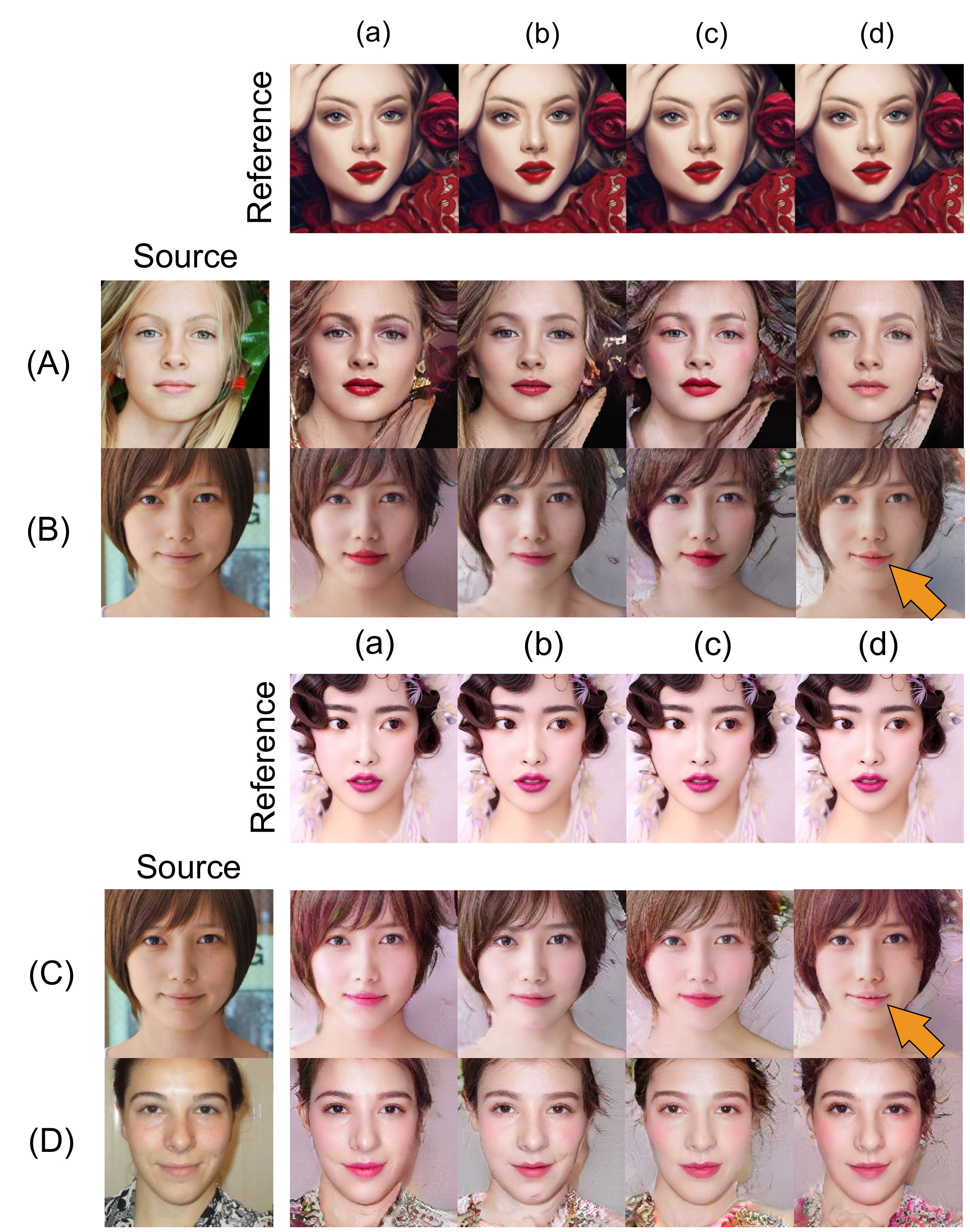}
    \caption{
    Ablation study results of perceptual makeup loss.
    We show the case where we use (a) a style encoder or (b) VGG16~\cite{simonyan2014deep} for feature extraction to compute a perceptual makeup loss.
    We also show cases of (c) where we had a makeup loss~\cite{beauty_gan} instead of a perceptual makeup loss and (d), where we had no makeup losses.
    }
  \label{fig:comparison_ablation_makeup_loss}
\end{figure}

\textbf{Perceptual Makeup Loss}
Figure~\ref{fig:comparison_ablation_makeup_loss} shows the effectiveness of our proposed perceptual makeup loss.
We show the case in which we used (a) a style encoder and (b) VGG16~\cite{simonyan2014deep} for feature extraction to compute perceptual makeup loss.
Moreover, we show cases in which (c) we used makeup loss~\cite{beauty_gan} instead of a perceptual makeup loss, and (d) we used no makeup losses.
Parameters of VGG16 pretrained on the Imagenet dataset were fixed.
Compared with case (b), we verified whether calculating a perceptual makeup loss with a style encoder improves the quality of a makeup style.
Additionally, we verified whether perceptual makeup loss was more effective than those in cases (c) and (d).

In Figure~\ref{fig:comparison_ablation_makeup_loss}, we observe that, given the reference and the source image (A), method (a) succeeded in the highest-quality MT.
Additionally, we observe that, given a source image (B) and a corresponding reference image, method (a) could generate an image that did not have any artifacts in the background.
Similar observations can be made about methods (C) and (D).
The orange arrows indicate that method (D) failed to transfer the lip makeup.
Therefore, we can see that the MT fails if we do not guarantee constraints on the matching of the color histogram between the reference and generated images.
This experiment shows that the constraints of our perceptual makeup loss using a style encoder was the most efficient way to transfer makeup.

\begin{figure}[t]
    \centering
    \includegraphics[width=\linewidth,bb=0 0 720 567]{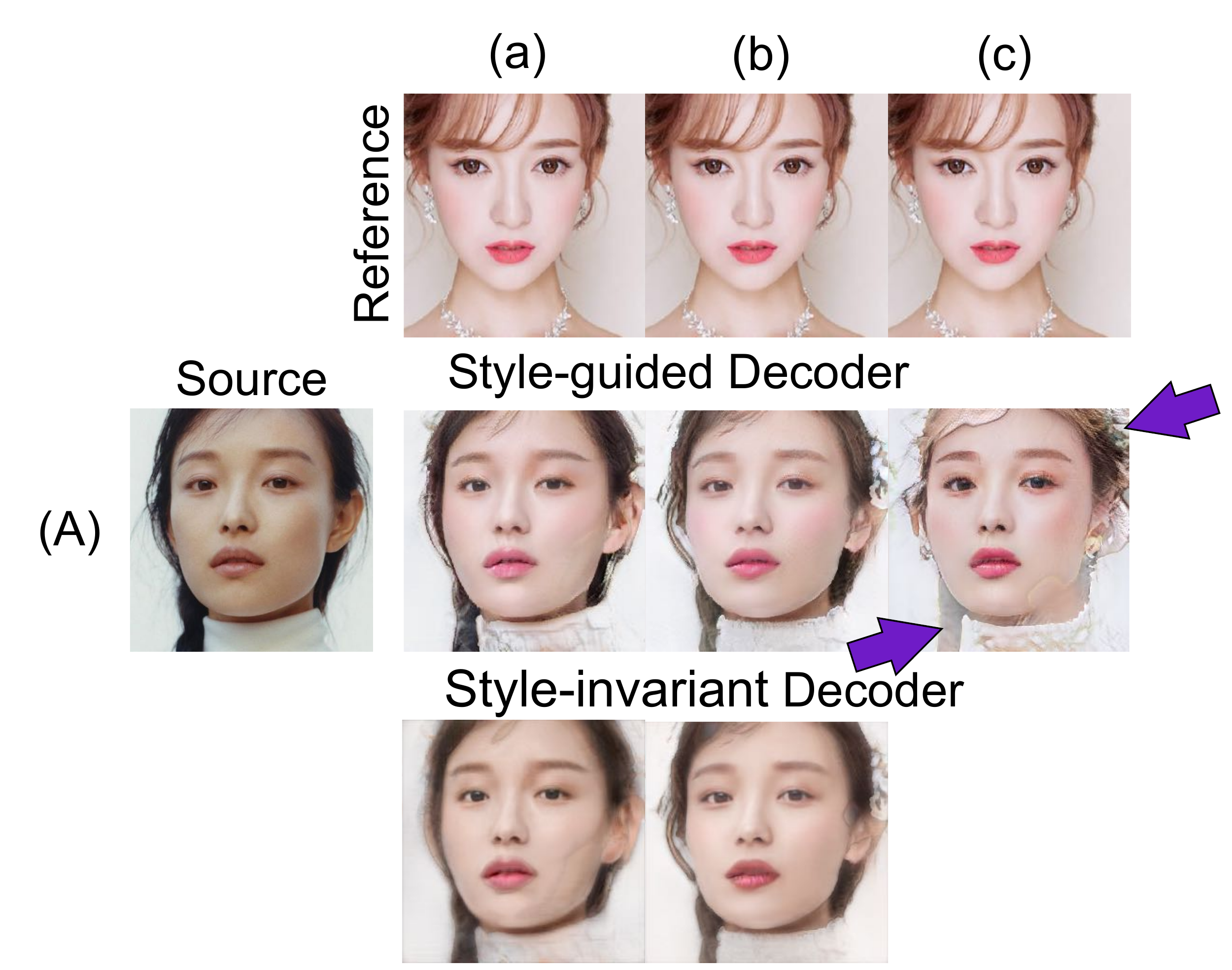}
    \caption{
    Ablation study results of a style-invariant decoder.
    To train the SLGAN, we present cases in which we used (a) an L2 norm and (b) an L1 norm in Eq.(\ref{loss:new_guide}), and (c) no style-invariant decoder.
    }
  \label{fig:comparison_ablation_guide_decoder}
\end{figure}

\textbf{Style-invariant Decoder}
Figure~\ref{fig:comparison_ablation_guide_decoder} shows the effectiveness of our proposed style-invariant decoder.
To train the SLGAN, we show the cases in which we used (a) an L2 norm, (b) an L1 norm in Eq.(\ref{loss:new_guide}), and (c) no style-invariant decoder.
In case (a), we observe that, given the source image (A), the style-guided decoder transferred the makeup.
However, in cases (b) and (c), the images generated by both style-guided and -invariant decoders left artifacts in the hair and backgrounds.
This is because the style encoder failed to extract only makeup features, because the L1 norm had a stronger shape-keeping constraint compared with that of the L2 norm.
Additionally, the purple arrows indicate that the generated image of (c) had an identity-shift in the hair.
Thus, we can see that our framework requires a style-invariant decoder.
Therefore, this experiment showed that the best approach is to give a reference image with only facial components and used an L2 norm in Eq.(\ref{loss:new_guide}) to train our network.

\subsection{Interpolation of Style Codes}\label{subsec:interpolation}
\begin{figure}[t]
    \centering
    \includegraphics[width=\linewidth,bb=0 0 994 316]{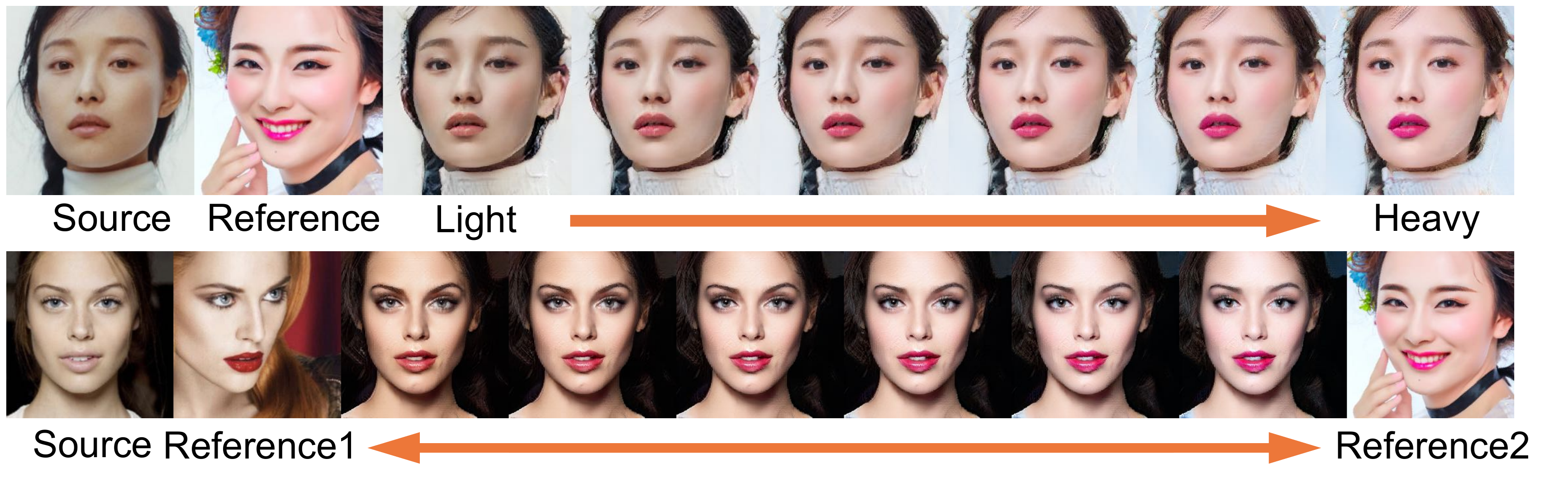}
    \caption{
    Interpolation results from light-to-heavy makeup generated by our latent-guided SLGAN with a single reference image (first row) and between the two reference images generated by our style-guided SLGAN (second row).
    }S
  \label{fig:makeup_transfer_interpolation}
\end{figure}

\begin{figure}[t]
    \centering
    \includegraphics[width=\linewidth,bb=0 0 1022 755]{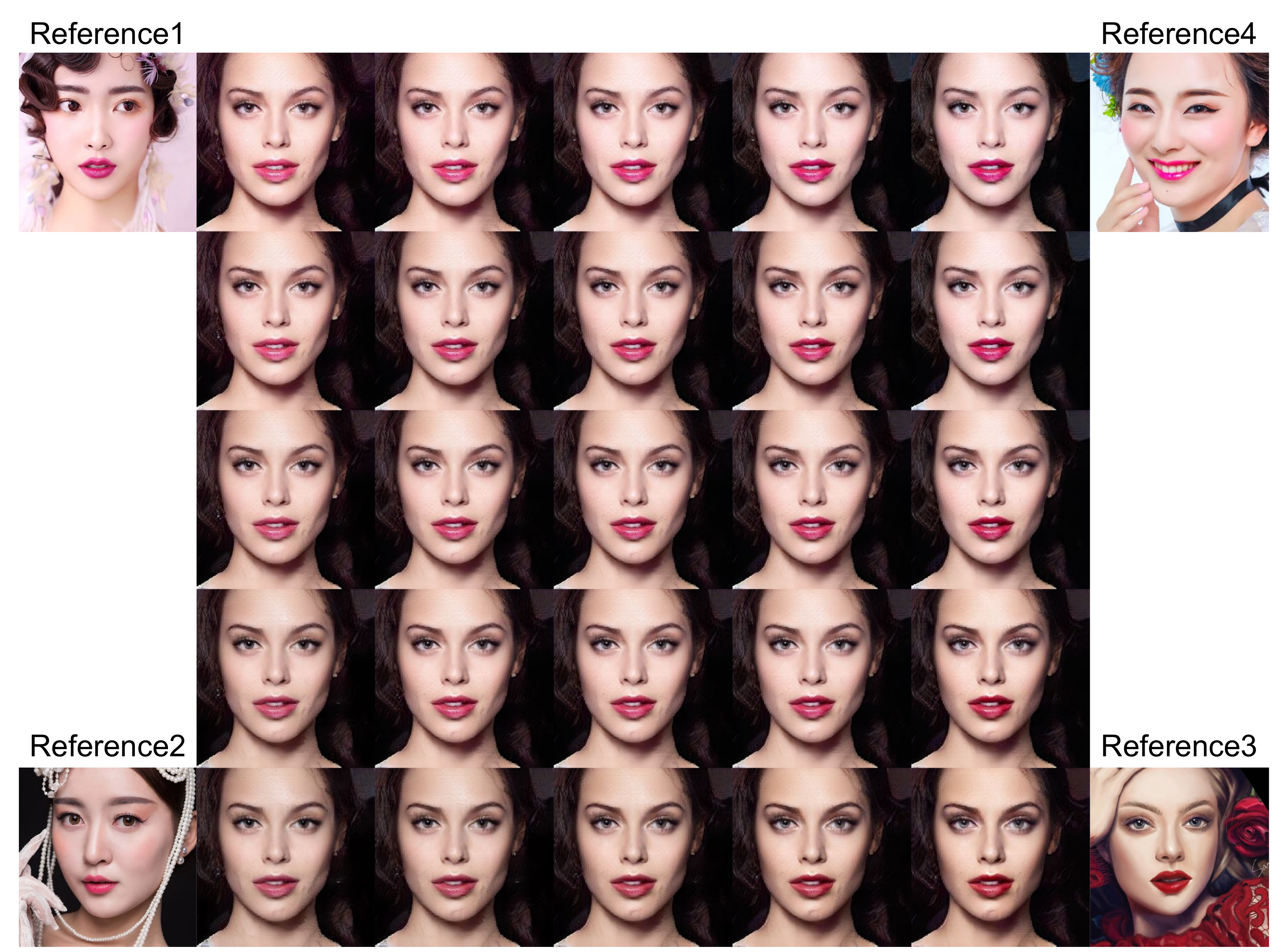}
    \caption{
    Interpolation results from among more than two references.
    }
  \label{fig:makeup_transfer_interpolation_fourth}
\end{figure}

It is important to make color adjustments to provide the best makeup suggestion to the user.
Figure~\ref{fig:makeup_transfer_interpolation} shows the MT interpolations generated by our proposed method.
We show the results from light-to-heavy makeup generated by our latent-guided SLGAN with a single reference image (first row) and between the two reference images generated by our style-guided SLGAN (secondrow).
Our proposed SLGAN interpolated not only a single-source image from a single-reference image, but it also interpolated a single-source image from multiple references.
Additionally, as shown in Figure~\ref{fig:makeup_transfer_interpolation_fourth}, our style-guided SLGAN interpolated between a set of $K$ reference images $I_{r}^{Y}{}_{1}, I_{r}^{Y}{}_{2},..., I_{r}^{Y}{}_{K}$, with corresponding weights $w_{1}, w_{2},..., w_{K}$, such that $\sum_{i=1}^{K} w_{i}=1$.
Our framework, therefore, has both style- and latent-guided architectures, and it enables the user to adjust makeup styles to find a desirable result.

\section{Conclusion}
    As a novel generation method, we proposed SLGAN, which, to our knowledge, is the first to apply a style- and latent-guided framework for MT and MR.
    Owing to its advantageous architecture, SLGAN generated more realistic images and performed better or comparable to state-of-the-art methods.
    Additionally, SLGAN produced interpolated generations using a reference image or latent code.
    This interpolation is beneficial for users who wish to find an optimal makeup configuration virtually.
    Furthermore, our novel perceptual makeup loss enables our framework to adequately transfer makeup styles, as shown in Figure~\ref{fig:comparison_ablation_makeup_loss}.
    Our novel style-invariant decoder further enabled our framework to avoid the identity-shifting problem by computing the Euclidean distance between the outputs of the decoder and the style-guided decoder, as shown in Figure~\ref{fig:comparison_ablation_guide_decoder}.
    In the experiments, our SLGAN performed better or comparably to state-of-the-art methods, and it maintained the unique ability to interpolate the MT and MR results.

\begin{acks}
We would like to thank JeongHun Baek, Koki Tsubota, Naoto Inoue, and Qing Yu for their insightful feedback on our project.
Daichi Horita was supported by Toyota and Dwango Scholarship.
This work was supported by JSPS KAKENHI Grant Number xxxx.
\end{acks}

\bibliographystyle{ACM-Reference-Format}
\bibliography{sample-base}

\end{document}